%% file: manuscript.tex
\documentclass[twocolumn]{svjour3}          
\smartqed  
\usepackage{graphicx}
\usepackage{natbib}
\usepackage{times}
\usepackage{epsfig}
\usepackage{amsmath}
\usepackage{amssymb}
\usepackage{lineno}
\usepackage{caption}
\usepackage{lipsum} 
\usepackage{booktabs}
\usepackage{enumitem}
\usepackage{adjustbox}
\usepackage{tikz}
\usepackage{pifont}
\usepackage{subfigure}

\include{macro}

\usepackage[pagebackref=true,breaklinks=true,letterpaper=true,colorlinks,bookmarks=false,citecolor=blue,linkcolor=blue]{hyperref}
\def\checkmark{\tikz\fill[scale=0.4](0,.35) -- (.25,0) -- (1,.7) -- (.25,.15) -- cycle;} 
\newcommand{\xmark}{\ding{55}}%
\begin{document}

\title{Perspectives and Prospects on Transformer Architecture for\\ Cross-Modal Tasks with Language and Vision
}


\author{
        Andrew Shin         \and
        Masato Ishii        \and
        Takuya Narihira 
}


\institute{
           Andrew Shin \at
              Sony Corporation\\
              \email{andrew.shin@sony.com}           
           \and
           Masato Ishii \at
              Sony Corporation\\
              \email{masato.a.ishii@sony.com}           
           \and
           Takuya Narihira \at
              Sony Corporation\\
            \email{takuya.narihira@sony.com}
}

\date{Received: date / Accepted: date}

\maketitle
\begin{abstract}
 Transformer architectures have brought about fundamental changes to computational linguistic field, which had been dominated by recurrent neural networks for many years. Its success also implies drastic changes in cross-modal tasks with language and vision, and many researchers have already tackled the issue. In this paper, we review some of the most critical milestones in the field, as well as overall trends on how transformer architecture has been incorporated into visuolinguistic cross-modal tasks. Furthermore, we discuss its current limitations and speculate upon some of the prospects that we find imminent.
\end{abstract}

\section{Introduction}
\label{sec:1}

\begin{figure*}[h]
  \centering
  \includegraphics[width=0.99\linewidth]{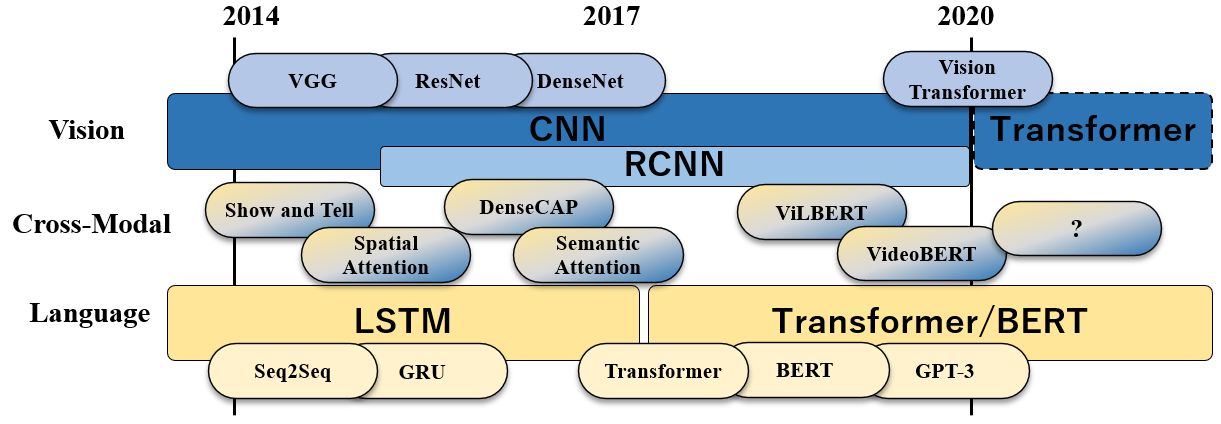}
  \caption{A brief timeline showing important models in vision (upper) and language (lower) domains respectively, and how they have affected approaches in cross-modal domains (middle).}
  \label{fig:overall}
  \vspace{-5mm}
\end{figure*}

Ever since the advent of deep learning revolution, the de-facto standard for cross-modal tasks with language and vision has been to use convolutional neural networks (CNN) (\cite{lecun-gradientbased-learning-applied-1998, NIPS2012_c399862d}) to extract features from visual domain, with VGG (\cite{simonyan2015deep}) or ResNet (\cite{7780459}) being the most frequently used CNN architectures, while employing recurrent neural networks (RNN) (\cite{Elman90findingstructure}), such as long short-term memory (LSTM) (\cite{HochSchm97}) or gated recurrent unit (GRU) (\cite{cho-etal-2014-learning}), to learn the language representation. While a plethora of variations exist as to specific ways to extract features and how to blend them into common embedding space, the fundamental pipeline has almost invariably been restricted to the combination of CNN and RNN. 

This steadfast landscape started to change with the introduction of transformer architecture (\cite{NIPS2017_3f5ee243}). Transformer has first demonstrated its capacity in natural language processing (NLP), achieving state-of-the-art performances in countless number of NLP tasks (\cite{peters-etal-2018-deep, dai-etal-2019-transformer, NEURIPS2019_dc6a7e65}), and has rapidly replaced recurrent neural networks. Its application has also been expanded to speech recognition domain (\cite{8462506,Wang_2020}). While a variety of transformer-based language models exist, 
BERT (\cite{devlin-etal-2019-bert}) has particularly gained wide attention not only with its performance, but also with its unique approach to pre-training and adaptability to downstream tasks. A GPT-line of works  (\cite{gpt1,Radford2019, NEURIPS2020_1457c0d6}) have also demonstrated that pre-training with a very large corpus and fine-tuning the model to a target task can outperform conventional models by a large margin. In particular, GPT-3 shows that pre-training with extremely large amount of corpus and parameters easily extends to high performance in few-shot learning tasks without performing any fine-tuning.

Success of transformer architecture in language domain has naturally led to its further application in cross-modal tasks involving language and vision. ViLBERT (\cite{NEURIPS2019_c74d97b0}) was one of the first models to demonstrate that pre-training objectives of BERT can be extended to cross-modal learning, and that it obtains state-of-the-art or comparable performances to models based on conventional CNN-RNN approach. Many other models followed with similar approach, and now the assumption that pre-training with a large amount of data leads to superior performance holds true for cross-modal domain as well. 

Yet, many important arguments can be raised with regards to the limitations and prospects of transformer-based cross-modal models. For example, while most models require that images or videos be tokenized and serialized in some way, they still fundamentally rely on CNN-based models to extract features for each visual token. It also remains arguable whether transformers learn fundamentally superior embeddings or whether its performance is simply due to a large amount of computations and data, bringing up the issue of computational efficiency. Furthermore, blending the transformer architecture with generative models is an open challenge, which is just beginning to be explored. 

Figure~\ref{fig:overall} shows a brief timeline over the past few years with important models in vision and language domains respectively, and how they have influenced the approaches for cross-modal tasks. Looking back at the important milestones and current trends, we project that transformer architecture may take over the vision representation part of cross-modal tasks, as its performance in vision domain has been shown to be comparable to CNNs (\cite{dosovitskiy2020image}). More about this will be discussed in Sec~\ref{sec:4} and Sec~\ref{sec:5}.

This paper attempts to review some of the representative works on transformer-based cross-modal models, with emphasis on pre-training schemes, discussing the common characteristics in various models along with their differences. We also investigate models of novel and promising direction, such as transformer-based vision representation, or image generation from text. By doing so, our goal is to acquire up-to-date insights as to various aspects of cross-modal learning, as well as prospects on how they may continue to influence deep learning field.

The rest of the paper is organized as following: we first review the conventional visuolinguistic cross-modal tasks and recent benchmark tasks, along with frequently used evaluation metrics in Sec~\ref{sec:2}. We also review the elementary transformer architecture and its variations, mostly restricted to language domain, centered around BERT in Sec~\ref{sec:transformer}. We then introduce and inspect the recent models on cross-modal tasks that employ transformer architecture in Sec~\ref{sec:3}, focusing on their architectural modifications and pre-training schemes. While the works introduced in Sec~\ref{sec:3} mostly employ transformer architecture to acquire linguistic representation, Sec~\ref{sec:4} introduces works that examine representing vision with transformer architecture, suggesting the possibility of replacing convolutional neural networks. Sec~\ref{sec:5} discusses some of the prospects on transformer architecture for cross-modal tasks. Finally, we conclude the paper in Sec~\ref{sec:6}, by summarizing the important points raised in the paper, while also discussing the current limitations and future works.

\section{Preliminaries I: Visuolinguistic Tasks}
\label{sec:2}

In this section and Sec~\ref{sec:transformer}, we briefly go over the preliminaries necessary to understand the topics and implications put forward by this paper. We first review some of the representative tasks in visuolinguistic domain, namely image captioning and visual question answering, with commonly used approaches. We also visit other important tasks, some of which have become benchmark tasks for transformer-based cross-modal tasks. We then go over evaluation metrics for these tasks.

.

\subsection{Classical Tasks}
\label{sec:2.1}

Prior to deep learning era, early models frequently tackled visuolinguistic cross-modal tasks with template-based model (\cite{barbu2012video, elliott-keller-2013-image, ushiku}) or ranking and retrieval models (\cite{6d7effd016974b2b88c071353ad5db9b, NIPS2011_5dd9db5e, journals/jair/HodoshYH13}). With the advent of deep learning, however, the mainstream paradigm for tackling cross-modal tasks has rapidly shifted towards the approach incorporating convolutional and recurrent neural networks.

Image captioning and
visual question answering (VQA) (\cite{Antol_2015_ICCV, balanced_binary_vqa, balanced_vqa_v2}) have conventionally been considered two of the most representative examples of cross-modal tasks involving language and vision. In image captioning, the model is trained with pairs of image and the captions describing that image, and learns to generate descriptive captions for unseen images. While early works relied on straightforward combination of CNN and LSTM, (\cite{7298935, DBLP:journals/corr/KarpathyF14}), advanced models appeared with dense localization (\cite{densecap}), semantic attention (\cite{pmlr-v37-xuc15, you2016image, zhou2016watch}), to name a few. Some works (\cite{Dai_2017_ICCV}) went steps further to incorporate generative adversarial networks (GANs) (\cite{NIPS2014_5ca3e9b1}) for image captioning, in which GANs are used to predict whether the captioning is natural.

VQA is a task in which a question in natural language is asked about an image, and the model is asked to provide an answer to the question. While many variations of VQA task exist, VQA as a benchmark task usually refer to the one based on MS COCO (\cite{lin2014microsoft}) with roughly 0.6M questions, each of which comes with 10 crowd-sourced answers. For VQA, most approaches can be categorized by 4 primary components; image representation, text representation, common embedding scheme, and attention mechanism. Image representation mostly relied on CNNs, often employing region detection models, while text representations may rely on RNN-line of models, Skipthoughts (\cite{kiros2015skipthought}), or more classical models, such as word2vec (\cite{NIPS2013_9aa42b31}) or Glove (\cite{pennington-etal-2014-glove}). While cross-modal embedding scheme may be simple concatenation or element-wise addition and multiplications, more sophisticated schemes have also been proposed, such as compact bilinear pooling (\cite{fukui-etal-2016-multimodal}), low-rank bilinear pooling (\cite{kim2017hadamard}), or cross-modal tucker fusion (\cite{benyounes2017mutan}). Various attention mechanisms (\cite{YangHGDS15, NIPS2016_9dcb88e0}) have also been proposed, and have demonstrated its effectiveness.\footnote{Note that attention mechanism referred to here, while nearly identical in its motivation, deviates from the same term employed in transformer architecture, and must not be confused with the works that do employ transformer architecture to tackle the same task, which will be introduced in Sec.~\ref{sec:3}}

Image captioning and VQA have also been extended to cross-modal domain involving language and video, as video captioning (\cite{Das2013ATF, gella-etal-2018-dataset}) and video QA (\cite{MovieQA}) respectively. As in image captioning and VQA, the models in video domain (\cite{DBLP:journals/corr/DonahueHGRVSD14, venugopalan-etal-2015-translating, DBLP:journals/corr/VenugopalanRDMD15}) still heavily relied on CNNs and RNNs for representing video and language.

\subsection{Benchmark Tasks}
While image captioning and visual question answering have been exemplary visuolinguistic cross-modal tasks, other important and intriguing variations have emerged. In particular, some of these novel variations, along with VQA task, have been consolidated as benchmark tasks for pre-trained cross-modal tasks. Interestingly, image captioning task, despite being a representative visuolinguistic task, has only rarely been tackled with recent transformer-based models. We conjecture that it may be attributable to the fact that image captioning task employs a different array of evaluation metrics, and makes it less intuitive to compare the performance along with other simple accuracy-based tasks, as we shall see in Sec.~\ref{sec:metric}. However, it is still being actively examined in video and language field, and is likely to remain an important indicator of cross-modal models' performances.

One of the frequently visited benchmark tasks is visual commonsense reasoning (VCR) (\cite{Zellers_2019_CVPR}), which extends VQA by not only asking a question about a referred agent's action, but also asking why the model chose that answer. This notably attempts to accomplish the logical inference made by humans. Three subtasks with varying difficulties exist within the task, where 1) the question is given and the model predicts the answer (Q$\rightarrow$A), or 2) the question and the answer are given and the model predicts the reason or intention (QA$\rightarrow$R), or 3) the question is given and the model predicts both the answer and the reason for that answer (Q$\rightarrow$AR). VCR dataset contains roughly 290k multiple choice questions accompanied by bounding boxes and semantic masks to indicate which agent the question or the answer candidate refers to.

Natural language for visual reasoning (NLVR) (\cite{suhr-etal-2019-corpus}) also requires an advanced understanding of vision and language, by judging whether a statement is true with regards to a juxtaposition of two images, based on the dataset consisting of roughly 100k human statements made about the pairs of images. The unique setting of the task in which there are two images to consider brings new challenges as to how to embed the images, \textit{e.g.}, whether they should be embedded separately or together, and how the decision should affect common embedding with text. RefCOCO (\cite{10.1007/978-3-319-46475-6_5}) is also a frequently visited benchmark task, where the goal is to identify which object in the image is referred to by the linguistic statement, based on the dataset consisting of 140k referring expressions from ReferitGame (\cite{kazemzadeh-etal-2014-referitgame}).
Grounding referring expressions (\cite{kazemzadeh-etal-2014-referitgame}), caption-based image retrieval (\cite{young-etal-2014-image}), and embodied visual recognition (\cite{yang2019embodied}) are also noteworthy tasks involving vision and language. Finally, image generation from text is also an important yet largely unexplored axis in visuolinguistic task. So far, it has mostly been limited to certain target domains, such as birds or flowers (\cite{reed2016generative, Zhang_2018_CVPR}), but promising works have started to appear, as we will see in Sec.~\ref{sec:5}.

As we shall see in Sec~\ref{sec:video}, benchmark tasks appear to be less consolidated in video and language domain, although video captioning tasks with YouCook2 (\cite{DBLP:journals/corr/ZhouXC17}) and ActivityNet (\cite{7298698}) are frequently visited.

\subsection{Evaluation Metrics}
\label{sec:metric}
Evaluation metrics for visuolinguistic tasks can be categorized as those for classification task and those for generation tasks, such as image captioning. While the classification tasks mostly rely on straightforward accuracy, generation tasks usually employ multiple metrics for evaluation. BLEU (\cite{papineni-etal-2002-bleu}), originally proposed for machine translation, is one of the most frequently used evaluation metrics and computes the portion of n-grams in candidate caption that also appear in ground truth.
While BLEU places emphasis on precision, ROUGE (\cite{lin-2004-rouge}) takes more recall-oriented approach by counting the occurrences of exact match of n-grams.
While BLEU and ROUGE look for exact match of words, METEOR (\cite{banerjee-lavie-2005-meteor}) evaluates n-grams considering synonyms and stems, based on WordNet (\cite{Miller95wordnet:a}).
CIDEr (\cite{vedantam2015cider}), designed specifically for image captioning task, looks at the consensus between generated image caption and the reference caption, with assigning more weights on n-grams that appear specifically for the image and less weights on n-grams that appear frequently for all images. SPICE (\cite{spice2016}) is also a popular metric for image captioning task, where a set of tuples are extracted from the semantic parse graph of the sentences. All of these metrics are also widely used for video captioning task as well.

Most of the benchmark tasks apart from captioning evaluate the performance with straightforward accuracy, although accuracy might be defined slightly differently depending on the task. For example, VQA defines accuracy with respect to the number of humans that provided the same answer out of all ground truth answers available for the question. Also, GQA (\cite{Hudson_2019_CVPR}) proposes to supplement accuracy by proposing additional metrics such as consistency, validity and plausibility. While the straightforward nature of the accuracy metric may risk loss of accountability for more subtle aspects of performance, such as the model's ability to yield the second best answer when not giving the correct answer, it is usually reported over a set of frequently visited tasks, and such aggregated evaluations usually appear to retain mutual agreement to a fair extent. Also, when necessary, accuracy is measured at different recall levels, as in image retrieval task shown in Table ~\ref{table:performance}.

\section{Preliminaries II: Transformer-Based models}
\label{sec:transformer}

In this section, we first describe transformer architecture by looking at multi-heads self-attention mechanism. We then introduce BERT, which is a transformer-based model and has become a crucial component in recent surge of transformer-based cross-modal models, with emphasis on its unique pre-training objectives. We also introduce pre-training objectives from models other than BERT for reference.

\subsection{Transformer Architecture}


\cite{NIPS2017_3f5ee243} proposed transformer architecture, and demonstrated that it outperforms the then-dominant RNN or CNN-based approaches on several sequential transduction tasks, such as machine translation. Transformer comprises an encoder part and a decoder part, and both of them consist of a series of self-attention based modules. Differently from CNN and RNN, it adopts self-attention as a basic operation in the model instead of convolution or memory gating, which leads to obtaining suitable properties for handling sequential data as will be shown later.

In a self-attention process, each input vector is first transformed into three vectors called \textit{query}, \textit{key}, and \textit{value}. The output is computed as a weighted sum of the values, where the weight of each value is assigned according to the similarity between the query and the corresponding key. Let $Q$, $K$, and $V$ be matrices that contain all queries, keys, and values extracted from the given input vectors, respectively. The self-attention process can then be formulated as
\begin{equation}
    \label{eqn:warping function}
    {\rm Attention}(Q,K,V) = {\rm softmax}(\frac{QK^T}{\sqrt{d_k}})V,
\end{equation}
where $d_k$ is a dimensionality of the key. 

To boost the flexibility of the self-attention process, the transformer adopted multi-head attention mechanism instead of the simple self-attention, as shown in Eq. (\ref{eqn:warping function}). In multi-head attention, several self-attention processes are conducted in parallel, and each output is integrated by concatenation followed by a linear projection to obtain the final output vector as shown in the following equations:
\begin{eqnarray}
    \label{eqn:warping function}
    {\rm MultiHead}(Q,K,V) = {\rm Concat}({\rm head}_1,..., {\rm head}_h)W^O, \\
    {\rm head}_i = {\rm Attention}(QW^Q_i,KW^K_i,VW^V_i),
\end{eqnarray}
where $W^O$ is a projection matrix to integrate the outputs of all attention processes. The queries, keys, and values for each attention process are computed by a linear projection of the original ones, and their projection matrices $W^Q_i$, $W^K_i$, and $W^V_i$ are jointly optimized with $W^O$ and other trainable parameters via training.

In the self-attention process, the order of the input vectors does not affect the resulting output vectors. For example, a model's prediction on a certain sentence does not depend on the order of words in the sentence, which should be inappropriate in many NLP tasks. To avoid this problem, position information of each input is encoded and added to the input embedded token before being fed to the transformer-based model. Specifically, the position information is encoded as following:
\begin{eqnarray}
    {\rm PE}_{(p, 2i)} = {\rm sin}(p/10000^{2i/d_{{\rm model}}}), \\
    {\rm PE}_{(p, 2i+1)} = {\rm cos}(p/10000^{2i/d_{{\rm model}}}),
\end{eqnarray}
where $p$ is the position of the target token in the input sequence, $d_{{\rm model}}$ and $i$ are the dimensionality of the embedded token and its index, respectively. As for positional embedding, other embedding methods have been proposed. For example, (\cite{wang2021on}) have shown that relative sinusoidal positional embedding outperforms absolute positional embedding in longer distances. Modified attention matrix (MAM) (\cite{DBLP:journals/corr/abs-2102-11090}) has also been proposed, where positional embedding is added as bias of attention map.

Some of the distinctive merits of transformer over previous models are modeling long-term dependency and its flexibility for parallel computing. For instance, RNNs generally find it difficult to preserve context as the distance between tokens becomes longer. Since they are executed in a sequential manner, it also becomes non-trivial to run them in parallel. Long-distance dependency is also challenging for CNNs, as it would require a proportionally large number of layers. On the other hand, attention mechanism can model dependency for words of any distance within the pre-defined sequence length, and as it does not require sequential order of the input, it is highly suitable for parallel, distributed computation.

\subsection{BERT}
\label{sec:2.2.2}
While a large number of variations on transformer exist, BERT (bidirectional encoder representations from transformers) (\cite{devlin-etal-2019-bert}) is particularly important in our topic, as its architecture has been employed by many cross-modal models, as well as its pre-training tasks that have been extended to account for cross-modal setting.
Following transformer, BERT applies layer normalization (\cite{ba2016layer}) on top of multi-head attention with residual connection, and applies feed-forward propagation with its unique choice of non-linear activation function, namely Gaussian error linear units (GELU) (\cite{DBLP:journals/corr/HendrycksG16}).

BERT is also known for its unique choice of pre-training tasks, namely masked language modeling and next-sentence prediction. In masked language modeling, tokens are randomly masked with 15\% probability, and the model is trained to predict those masked tokens. For next sentence prediction, two sentences are provided, where the second sentence may be actual succeeding sentence or a random sentence with 50\% probability, and the model is given a task of binary classification of whether the second sentence is an actual succeeding sentence. While the first objective of masked language modeling aims to learn token-level dependency, the motivation of the second task is to learn the inter-sentence relationship. Although (\cite{liu2019roberta}) has shown that the next sentence prediction turns out not to be crucial for the model's performance, these two objectives, along with the bidirectional architecture, characterize and differentiate BERT from other transformer-based language models.

For reference, we would like to highlight a few more notable pre-training objectives proposed by other models. XLNet (\cite{NEURIPS2019_dc6a7e65}) attempts to overcome some of the drawbacks of masked language modeling of BERT, \textit{e.g.,} the absence of masked tokens in real data, or its inability to model the joint probability with product rule used in autoregressive language modeling.
Specifically, they propose permutation language modeling, where the model maximizes likelihood of the input sequence over all possible permutations of the factorization order, thereby learning bidirectional context while retaining the merits of autoregressive models. MT-DNN (\cite{liu-etal-2019-multi}) performs multi-task learning, which leverages supervised data from multiple related tasks, such as single-sentence or pairwise text classification and relevance ranking, on top of language model pre-training. The paper claims that it prevents the model from overfitting to a specific task, with complementary effect to general language model pre-training. (\cite{Chang2020Pre-training}) proposes three pre-training objectives with emphasis on capturing granularity of semantics between the query and document, namely inverse cloze task to capture local semantic context, body first selection to capture global context within the document, and wiki link prediction to capture distant inter-article context. Their experiments suggest that the proposed pre-training objectives can perform significantly better than masked language modeling.

\section{Analysis of Cross-Modal Embeddings}
\label{sec:3}
We now look into cross-modal models that have employed transformer/BERT architectures. We first review their pre-training objectives, many of which have been directly inspired by BERT, albeit with notable variations, and speculate on how they relate to the models' performances. We also look at the models' attempts to construct network architectures that learn inter-modal dependency, along with how they deal with distinct modalities for mutual compatibility.

\begin{figure*}[h]
  \centering
  \includegraphics[width=0.99\linewidth]{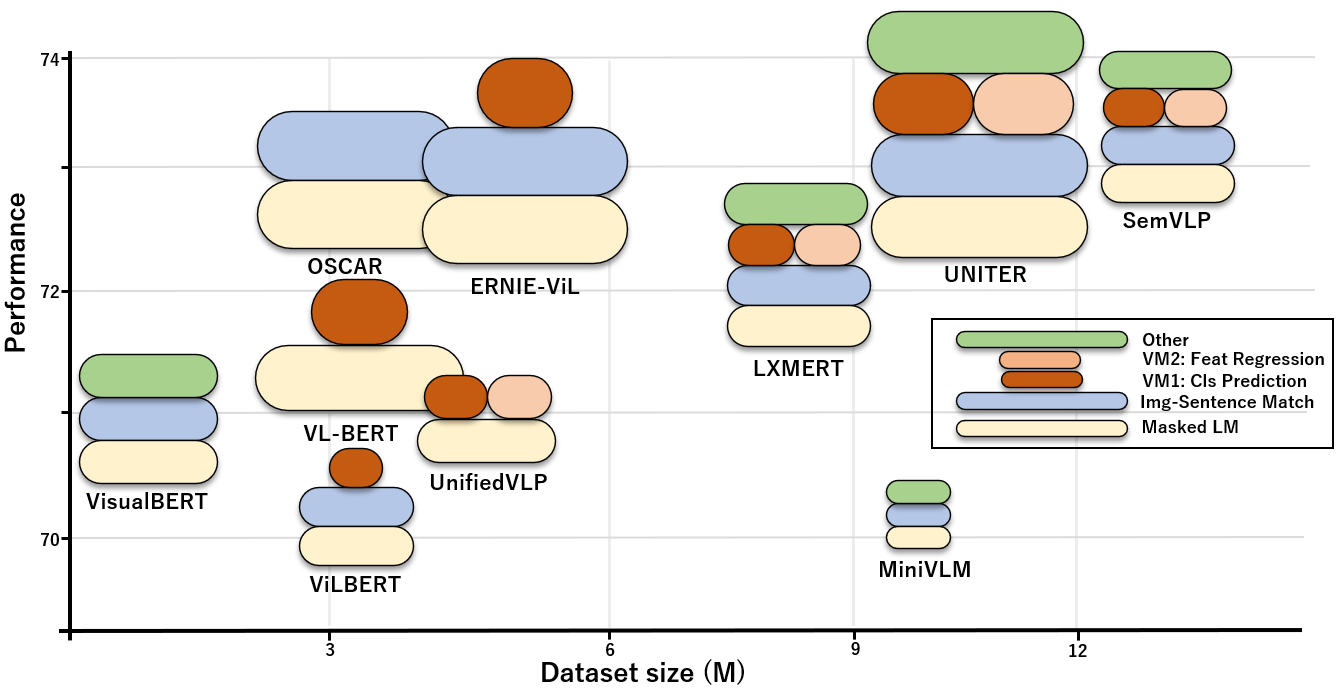}
  \caption{\small Visualization of representative cross-modal models in terms of dataset size used for pre-training, performance as measured by VQA task, the type of pre-training objectives employed, and the model size. The size of block for respective model corresponds to the size of the model, as approximated by the implementation details reported in respective paper. VM1 and VM2 refer to two distinct tasks of masked visual modeling with image regions.}
  \label{fig:performance}
\end{figure*}

\subsection{Pre-training Tasks}
\label{sec:3.1}

\begin{table*}
\begin{center}
\resizebox{\linewidth}{!}{
\begin{tabular}{l|ccccccccc}
\toprule
&VQA & GQA & \multicolumn{3}{c}{VCR} & RefCOCO &NLVR2& IR & ZS IR\\
& dev/std & dev/std & Q/A& QA/R & Q/AR & val/ testA/testB& test-P &R1/R5/R10 & R1/R5/R10 \\ 
\midrule
$\text{ViLBERT}_{\text{nopt}}$(\cite{NEURIPS2019_c74d97b0}) & 68.93/-- &-- & 69.26 & 71.01 & 49.48 & 68.61/75.97/58.44 &-- & 45.50/76.78/85.02 & 0.00/0.00/0.00 \\
ViLBERT & 70.55/70.92 & --& 73.3 & 74.6 & 54.8 & 72.34/78.52/62.61 &--& 58.20/84.90/91.52 & 31.86/61.12/72.80\\
B2T2(\cite{alberti-etal-2019-fusion}) &-- &-- & 72.6 & 75.7 & 55.0 & --& -- & -- & --\\
$\text{B2T2}_{\text{ensemble}}$ & --&-- & 74.0 & 77.1 & 57.1 & --& -- & -- & --\\
$\text{VL-BERT}_{\text{nopt}}$(\cite{Su2020VL-BERT:})&69.58/--&-- &--&--&--&66.03/71.87/56.13& -- & -- & --\\
VL-BERT &71.16/-- & --&--&--&--& 71.60/77.72/60.99& -- & -- & -- \\
$\text{VL-BERT}_{\text{large}}$ &71.79/72.22& --& 75.8 & 78.4 & 59.7 & 72.59/78.57/62.30 & -- & -- & --\\
LXMERT(\cite{tan-bansal-2019-lxmert}) & --/72.5 & 60.0/60.3&--&--&--&--&--&--&--  \\
$\text{Unicoder-VL}_{\text{zs}}$(\cite{Li_Duan_Fang_Gong_Jiang_2020}) & --&-- &--  &--  & --&--&--&48.4/76.0/85.2&-- \\
$\text{Unicoder-VL}_{\text{nopt}}$ &-- & --& -- &--  &-- &--&--&57.8/82.2/88.9&-- \\
Unicoder-VL &-- &-- & 73.4 & 74.4 & 54.9 &--&--&71.5/90.9/94.9&--\\
VisualBERT(\cite{li2019visualbert}) & 70.80/71.00 &--& 71.6 & 73.2 & 52.4 & --& 67.0&--&--\\
$\text{UNITER}_{\text{base}}$(\cite{chen2020uniter}) &72.70/72.91 & --& 75.0 & 77.2 & 58.2 & 75.31/81.30/65.58 & 77.85 & 72.52/92.36/96.08 & 66.16/88.40/92.94\\
$\text{UNITER}_{\text{large}}$ &73.82/74.02 &-- & 77.3 & 80.8 & 62.8 & \textbf{75.9}/\textbf{81.45}/\textbf{66.7} & 79.98 & \textbf{75.56}/\textbf{94.08}/\textbf{96.76} & \textbf{68.74}/\textbf{89.2}/\textbf{93.86}\\
MiniVLM(\cite{wang2020minivlm}) & 69.39/69.06 &-- &--&--&--&--& 73.93&--&--\\
$\text{PixelBERT}_{\text{r50}}$(\cite{huang2020pixelbert}) & 71.35/71.42 &--&-- &--&--&--& 72.4 & 59.8/85./91.6 &--\\
$\text{PixelBERT}_{\text{x152}}$ & 74.45/74.55 &--&-- &--&--&--& 77.2 & 71.5/92.1/95.8&--\\
$\text{OSCAR}_{\text{base}}$(\cite{li2020oscar}) & --/73.44 &--&-- &--&--&--& 78.36&--&--\\
$\text{OSCAR}_{\text{large}}$ & --/73.82 &61.58/61.62&-- &--&--&--& \textbf{80.37}&--&--\\
UnifiedVLP(\cite{zhou2019unified}) & 70.5/70.7&--&--&--&--&--&--&--&--\\
$\text{InterBERT}_{\text{nopt}}$(\cite{lin2021m6v0}) & --&-- & 63.6 & 63.1 & 40.3 &--&-- & 53.1/80.6/87.9&-- \\
InterBERT &-- &-- & 73.1 & 74.8 & 54.9&-- &-- & 61.9/87.1/92.7 & 49.2/77.6/86.0  \\
$\text{ERNIE-ViL}_{\text{base}}$(\cite{yu2020ernievil}) & 72.62/72.85&-- & 77.0 & 80.3 & 62.1 & 74.02/80.33/64.74 &-- & 74.44/92.72/95.94&--\\
$\text{ERNIE-ViL}_{\text{large}}$ & 73.78/73.96 &-- & \textbf{79.2} & \textbf{83.5} & \textbf{66.3} & 74.24/80.97/64.70 &-- & 75.10/93.42/96.26&--\\
$\text{DeVLBERT}_{\text{V}}$(\cite{Zhang_2020}) &-- &-- &--&--& --&-- & --&59.3/85.4/91.8 & 32.8/63.0/74.1 \\
$\text{DeVLBERT}_{\text{VL}}$ &-- &-- &--&--&-- &-- & --&60.3/86.7/92.2 & 34.9/65.5/77.0\\
$\text{DeVLBERT}_{\text{VLC}}$ & 71.1/71.5 &-- &--&--&-- &-- & --&61.6/87.1/92.6 & 36.0/67.1/78.3 \\
SemVLP(\cite{li2021semvlp}) & \textbf{74.52}/\textbf{74.68} & \textbf{62.87}/\textbf{63.62} & -- &--&--&-- /--/--& 79.55 & 74.10/92.43/96.12&-- \\
CAPT(\cite{luo2020capt}) & 72.78/73.03 & 60.48/60.93 &--&--&--&--& 75.13 &--&--\\
$\text{LAMP}_{\text{coco}}$(\cite{guo2020lamp}) & 70.85/71.0 & --&-- &--&--&--& 74.34 & --& 42.5/70.9/80.8 \\
$\text{LAMP}_{\text{coco+vg}}$ & 72.48/72.62 & --/61.05 &-- &--&--&--& 75.43 &--& 51.8/77.4/85.3 \\
(\cite{kervadec2019weak}) &-- & --/60.5 & --&--&--& --& 75.5&--&-- \\
ImageBERT(\cite{qi2020imagebert}) &-- &-- &-- &--&--&--&-- & 73.1/92.6/96.0 & 54.3/79.6/87.5\\
Human &-- & 89.3 & 91.0 & 93.0 & 85.0 &--&96.3 &--&--\\
\bottomrule
\end{tabular}
}
\caption{\small Comparison of performances of cross-modal models on various tasks. The subscript \textit{nopt} refers to the model without pre-training, \textit{zs} refers to zero-shot. Other subscripts follow respective papers. Human performances are compiled from respective papers for the tasks.} 
\label{table:performance}
\end{center}
\end{table*}

\begin{table*}
\begin{center}
\begin{tabular}{l|cccc}
\toprule
 & Masked LM & Img-Sentence Match & Masked Visual Modeling & Other \\ 
\midrule
ViLBERT & \checkmark & \checkmark & class prediction & -- \\
B2T2 & \checkmark & \checkmark & \xmark & -- \\
VL-BERT & \checkmark & \xmark & class prediction & -- \\ 
LXMERT & \checkmark & \checkmark & class prediction+feature regression & image QA \\
Unicoder-VL & \checkmark & \checkmark &  class prediction & -- \\ 
VisualBERT & \checkmark & \checkmark & \xmark & task-specific masked LM \\
UNITER & \checkmark & \checkmark & class prediction+feature regression & word region alignment \\
MiniVLM & \checkmark & \checkmark & \xmark & detection+captioning+tagging \\
PixelBERT & \checkmark & \checkmark & \xmark & pixel random sampling\\
OSCAR & \checkmark & \checkmark & \xmark & --\\
UnifiedVLP & \checkmark &\xmark & class prediction+feature regression & --\\ 
InterBERT & \checkmark (same word) & \checkmark (hard negative) & class prediction & --\\
ERNIE-ViL & \checkmark & \checkmark & class prediction & --\\ 
DeVLBERT & \checkmark & \checkmark & class prediction & --\\
SemVLP & \checkmark & \checkmark & class prediction+feature regression & image QA \\ 
CAPT& \checkmark & \checkmark & class prediction & contrastive\\
LAMP &\checkmark &\checkmark & class prediction+feature regression & --\\
(\cite{kervadec2019weak}) & \checkmark & \checkmark & class prediction & VQA\\
ImageBERT & \checkmark & \checkmark & class prediction+feature regression & --\\
\bottomrule
\end{tabular}
\caption{\small Pre-training tasks performed by various cross-modal models.} 
\label{table:pretraining}
\end{center}
\end{table*}


\begin{table*}
\begin{center}
\resizebox{\linewidth}{!}{
\begin{tabular}{l|cccccc}
\toprule
 & Dataset & Computation & Optimizer & Batch Size & BERT L/H/A & Cross-Modal Depth\\ 
 \midrule
ViLBERT & CC & 8 TitanX& Adam & 512 & 12/12/762 & 8\\
B2T2 & CC & -- & Adam &-- & 24/16/1024 & --\\
VL-BERT & CC & 16 V100 & Adam & 256 &24/16/1024&--\\
LXMERT & COCO/VG/VQA/GQA/VG-QA & 4 Titan Xp & Adam & 256 &12/12/768 & 9 \\
Unicoder-VL &CC/SBU & 4 V100 & Adam & 192 & 12/12/768&--\\
VisualBERT & COCO & 4 V100 & Adam &-- & 12/12/768&--\\
UNITER & COCO/VG/CC/SBU &-- &--&--&24/16/1024 & --\\
MiniVLM & COCO/CC/SBU/Flicker/VQA/GQA/VG-QA &-- &SGD & 512 & 12/--/386 & --\\
PixelBERT & COCO/VG &64 V100 &  SGD/AdamW & 4096&--&--\\
OSCAR & COCO/CC/SBU/Flicker/GQA & --& AdamW & 512 & 24/16/1024 & --\\
UnifiedVLP & CC & 8 V100 & AdamW & 512 & 12/12/768&--\\
InterBERT & CC/COCO/SBU & --& AdamW&--&--&--\\ 
ERNIE-ViL & CC/SBU & 8 V100 & Adam & 512 & 24/16/1024 & 6\\
DeVLBERT & CC & 8 V100 & --&--&--&--\\
SemVLP & COCO/VG/VQA/GQA/VG-QA/CC/SBU & 4 V100 & Adam & 256 & 12/12768 & 6\\
CAPT &-- &-- & Adam & --& 24/16/1024 & --\\
LAMP & COCO/VG & 2 Titan RTXs &-- & 256 & 12/12/768 & --\\
\cite{kervadec2019weak} & COCO/VG &4 V100 &  Adam & 512 & 12/12/768 & 5\\ 
ImageBERT & LAIT &  4 V100& Adam & 48&  12/12/768&--\\ 
\bottomrule
\end{tabular}
}
\caption{\small We report the setting for the largest model reported in respective papers. COCO refers to MS COCO (\cite{lin2014microsoft}), CC to Conceptual Captions (\cite{sharma-etal-2018-conceptual}), VG to Visual Genome (\cite{krishnavisualgenome}), VG-QA to Q\&A subset of Visual Genome, and SBU to SBU Captions (\cite{NIPS2011_5dd9db5e}). LAIT is collected by ImageBERT. BERT L/H/A refer to the number of layers, the number of attention heads, and hidden size, respectively. Cross-modal depth refers to the number of inter-modality layers, after embedding each modality separately. We report cross-modal depth only for two-stream models, since most single stream models follow the same network depth as their language embedding for cross-modal embedding.} 
\label{table:setting}
\end{center}
\vspace{-6mm}
\end{table*}

One of the primary arguments laid by the transformer-based language models, including BERT, is that large-scale pre-training is the key to their successful performance. It follows evidently that the choice of pre-training tasks is essential in obtaining a high-quality language model. Cross-modal models employing transformer architecture have generally followed the same presumption, placing a strong emphasis on the design of pre-training objectives, and the key challenge is consequently how to replicate large-scale pre-training with cross-modal setting. For example, as shown in Sec~\ref{sec:2.2.2}, BERT is known for its unique pre-training tasks of masked language modeling and next sentence classification, but since it is designed solely for language, it inevitably requires adjustments to be extended to cross-modal setting. While most models adopt those two pre-training tasks in modified ways, some models propose additional pre-training tasks designed specifically for cross-modal setting.

Masked language modeling, one of the two pre-training tasks of BERT, is used as is for linguistic input tokens in cross-modal models, almost without exception. One notable variation is InterBERT (\cite{lin2021m6v0}), which performs masked language modeling by replacing the masked word with continuous words, rather than random words as with most other models. B2T2 (\cite{alberti-etal-2019-fusion}) also extends the task by training language model while seeing the image. Another pre-raining task of BERT, next sentence prediction, is mostly converted to binary classification of whether input image and sentence are semantically matched. This also is performed by most cross-modal models, with few exceptions such as VL-BERT (\cite{Su2020VL-BERT:}) or UnifiedVLP (\cite{zhou2019unified}) that explicitly opt not to perform this pre-training task. InterBERT again makes a unique variation by explicitly involving matching with hard negative examples.

One of the key challenges for pre-training tasks among cross-modal embedding models is with the way they implement masked language modeling task for visual inputs, as it cannot be naively extended to vision domain in a straightforward manner, due to the non-sequential nature of vision. In fact, some models, such as B2T2 and VisualBERT (\cite{li2019visualbert}), simply opt not to perform any extended masked modeling task for visual inputs. On the other hand, many models propose novel ways to apply masked language modeling for visual tokens. For example, ViLBERT handles this challenge by proposing to mask the image regions extracted by Faster R-CNN (\cite{NIPS2015_14bfa6bb}), and trains the model to predict the class distribution of the region, with the class distribution output from Faster R-CNN as the ground truth. VL-BERT (\cite{Su2020VL-BERT:}) and Unicoder-VL (\cite{Li_Duan_Fang_Gong_Jiang_2020}) are also notable examples that perform masked visual modeling with class prediction. In particular, VL-BERT proposes masked RoI classification with linguistic clues, where a region of interest (for example, a \textit{cat}) is randomly masked out, and the model is trained to predict the category of the masked out RoI by solely relying on linguistic clues, such as  ``\textit{kitten drinking from bottle}.'' Here, the RoIs are obtained via Fast R-CNN (\cite{Girshick_2015_ICCV}). Such setting is derived from the concern that there may be cases where it is nearly impossible to identify a region when the region is masked. While VL-BERT opts not to perform image-sentence matching task, this unique setup intends to make up for the learning of cross-modal dependency.

While the models above relied on the prediction of class distribution, it has also been found beneficial to incorporate feature regression into masked visual modeling task. LXMERT (\cite{tan-bansal-2019-lxmert}) is one such example. On top of masked object classification,  where the label for the masked RoI should be classified based on other visual inputs and linguistic inputs, they also perform RoI-feature regression with L2 loss. UnifiedVLP also reports that combining class prediction and feature prediction improves performance. UNITER (\cite{chen2020uniter}) also performs both class prediction and feature regression, but they propose to add a third task for masked visual modeling, where they perform class prediction with KL divergence.

Some models further propose novel pre-training tasks that do not fall into any of the 3 categories above, which frequently depends on the target downstream task. For example, LXMERT, SemVLP (\cite{li2021semvlp}) and (\cite{kervadec2019weak}) propose image question answering task for pre-training, whereas PixelBERT (\cite{huang2020pixelbert}) employs pixel random sampling. (\cite{luo2020capt}) proposes contrastive pre-training, where they encourage the model to learn similar representations for sequences that share the same semantics by matching the original sequence and the corresponding corrupted sequence. 

While most models so far have relied on Fast/Faster R-CNN for region extraction, MiniVLM (\cite{wang2020minivlm}) employs a region extraction module inspired by EfficientDet (\cite{tan2020efficientdet}), and as such, they first attempt to enhance their visual features by performing large-scale image classification and object detection with Objects365 dataset (\cite{Shao_2019_ICCV}). Cross-modal representation is subsequently fine-tuned with captioning and tagging, where captions and tags are obtained by existing models.

\subsubsection{Analysis of Performance}

Table~\ref{table:performance} compares the performance of various models on representative cross-modal tasks. Note that the models are trained with different size of datasets with varying size of model sizes, so direct comparison of performance does not necessarily imply superiority of certain models over others. Table~\ref{table:pretraining} compares pre-training tasks performed by each model, with 4 rough categories for pre-training tasks. Table~\ref{table:setting} compares settings of the models, ranging from datasets used for pre-training to their implementation details. Figure~\ref{fig:performance} visualizes representative models in terms of the dataset size used for pre-training, their performance in VQA task, the types of pre-training objectives performed, and the respective model size. While it remains arguable whether we can attribute a certain aspect to the respective model's performance, the models with higher performance have one or more of the following characteristics in common; pre-trained on a relatively large amount of data, larger model size, and more pre-training objectives. While the observations are quite generic, note that they are consistent with recent trend of pre-training a model with a very large number of parameters with large-scale dataset, as affirmed by GPT-3.

Table ~\ref{table:dataset} shows some of the frequently used datasets for both images and videos, along with the description of their contents, annotation types, and sizes. Although it would be difficult to generalize from a small group of models described so far, as it is easy to be overfitting, further observations can be made with respect to dataset and pre-training objectives that may likely be of reference. It must first be noted that all datasets with images are at the order of millions, once again affirming the indispensable role played by pre-training with large-scale dataset. In fact, even models of relatively small sizes can attain fair performance when pre-trained with large-scale dataset, as is the case with mini-VLM. Pre-training with large-scale dataset also appears to overcome the noisiness of labels obtained via weak, web supervision. For example, ERNIE-ViL is pre-trained with Conceptual Captions and SBU Captions, both of which are "\textit{webly}" supervised, yet demonstrates fairly high performance throughout the tasks. To be fair, however, under similar settings, pre-training with manually annotated datasets appears to provide an edge over pre-training with datasets with web supervision, particularly in terms of the size of the dataset necessary to obtain the desirable level of performance, which makes sense considering the inevitable presence of noise in web supervision. Higher efficiency of human annotation in terms of the required dataset size is well-illustrated by VisualBERT and a variation of LAMP, both of which are trained only with MS COCO, displaying more reliable performance than ViLBERT trained with Conceptual Captions, although Conceptual Captions is more than 5 times larger than MS COCO. As noted above, however, it should be highlighted again that this is equivalent to saying, alternatively, that pre-training with web supervision can demonstrate comparable performance to human annotations when given a sufficient amount of data, as is the case with models like ERNIE-ViL. As we shall see in Sec.~\ref{sec:video}, a similar tendency can also be observed with video and language models. For example, VideoBERT (\cite{Sun_2019_ICCV}), which is pre-trained with large-scale YouTube videos with annotations obtained by YouTube video annotation system and uses human-annotated YouCook2 (\cite{DBLP:journals/corr/ZhouXC17}) for evaluation only, falls below the models directly pre-trained with YouCook2, such as UniVL (\cite{luo2020univl}).

It may also be inferred that pre-training with dataset designed for the same or similar task as the target task helps boost performance in that target task. LXMERT, OSCAR, and SemVLP, which are trained with at least one dataset designed for image QA task, all demonstrate higher performance on VQA task, as shown in Fiugre ~\ref{fig:performance}. Similarly, in terms of correlation between pre-training objectives and performance on specific tasks, it appears beneficial to perform similar type of pre-training task for a certain target downstream task. LXMERT and SemVLP are again notable examples, as they propose unique pre-training objective of image QA and indeed achieve better performances on VQA task. MiniVLM also performs additional pre-training objective of image captioning task to demonstrate reliable performance on image captioning task. Furthermore, video and language models pre-trained with YouCook2 annotations can be considered a similar case.

\begin{table*}
\begin{center}
\begin{tabular}{l|ccc}
\toprule
  Dataset & Contents & Annotation & Size \\ 
 \midrule
Conceptual Captions (\cite{sharma-etal-2018-conceptual}) & image + scene-level description & web supervision & 3.3M\\
MS COCO (\cite{lin2014microsoft}) & image + scene-level description & crowd-sourced & 0.6M\\
Visual Genome (\cite{krishnavisualgenome}) & image + region descriptions & crowd-sourced & 5.4M \\
Visual Genome - Q\&A (VG-QA) & image + question and answer & crowd-sourced & 1.7M\\
SBU (\cite{NIPS2011_5dd9db5e}) &image + scene-level description& web supervision & 1M\\
Flickr 30k (\cite{young-etal-2014-image}) &image + scene-level description& crowd-sourced & 0.2M \\
VQA (\cite{Antol_2015_ICCV})& image + question and answer & crowd-sourced & 0.6M\\
GQA (\cite{Hudson_2019_CVPR})& image + question and answer &generated & 22M \\
\hline
YouCook2 (\cite{DBLP:journals/corr/ZhouXC17})& video + temporal annotations & manual annotation & 15k \\
HowTo100M (\cite{miech19howto100m})& videos + descriptions & audio transcription& 136M\\
ActivityNet Captions (\cite{7298698})& videos + temporal annotations & crowd-sourced & 72k\\
MSR-VTT (\cite{xu2016msr-vtt})& videos + descriptions & crowd-sourced & 200k\\
CrossTask (\cite{Zhukov_2019_CVPR})& videos + task descriptins &manual annotation & 4.7k\\
CMU-MOSI (\cite{zadeh_multimodal_2016})& videos + sentence-level sentiment annotations & crowd-surce & 2k\\
\bottomrule
\end{tabular}
\caption{\small Comparison of frequently used datasets for pre-training visuolinguistic models. Web supervision refers to captions obtained by crawling the tags or descriptions for corresponding images uploaded on the web pages. Generated refers to synthetic sentences generated in a rule-based manner. Crowd-sourced and manual annotation both refer to human annotation, but differ in that the former refers specifically to external workforce. Size refers to the number of descriptions or questions available.} 
\label{table:dataset}
\end{center}
\end{table*}

\begin{table*}
\begin{center}
\begin{tabular}{l|cccccccc}
\toprule
& \multicolumn{4}{c}{YouCook2}&\multicolumn{4}{c}{ActivityNet} \\
Method & BLEU-4 & METEOR & ROUGE-L & CIDEr & BLEU-4 & METEOR & ROUGE-L & CIDEr\\ 
\midrule
Masked Transformer (\cite{Zhou_2018_CVPR})& 1.42 & 11.20 & -- & -- & 2.23 & 9.56&--&-- \\
VideoBERT (\cite{Sun_2019_ICCV})& 4.33 & 11.94 & 28.80 & 0.55&--&--&--&--\\
CBT (\cite{sun2020learning})& 5.12 & 12.97 & 30.44 & 0.64 &--&--&--&--\\
COOT (\cite{ging2020coot})& 11.30 & 19.85 & 37.94 & 0.57 & 10.85 & 15.99 & 31.45 & 0.28\\
UniVL (\cite{luo2020univl})& 17.35 & 22.35 & 46.52 & 1.81 & -- & -- & -- & --\\
ActBERT (\cite{Zhu_2020_CVPR})& 5.41 & 13.30 & 30.56 & 0.65 & -- & -- & -- & --\\
E2vid (\cite{huang-etal-2020-multimodal})& 12.04 & 18.32 & 39.03 & 1.23 & -- & -- & -- & --\\

\bottomrule
\end{tabular}
\caption{\small Performances of transformer-based cross-modal models on video captioning for YouCook2 (\cite{DBLP:journals/corr/ZhouXC17} and ActivityNet (\cite{7298698}). Metrics used include BLEU (\cite{papineni-etal-2002-bleu}), METEOR (\cite{banerjee-lavie-2005-meteor}), ROUGE (\cite{lin-2004-rouge}), and CIDER (\cite{vedantam2015cider}).}
\label{table:video_caption}
\end{center}
\vspace{-4mm}
\end{table*}

\subsubsection{Video and Language}
\label{sec:video}
Extending vision domain of cross-modal tasks from images to videos naturally brings new challenges. For example, a truly reliable supervision would require annotations at frame-level, which necessitates prohibitive amount of manual labour. In fact, existing large-scale video datasets, such as YouTube-8M (\cite{DBLP:journals/corr/Abu-El-HaijaKLN16}) or  Sports-1M (\cite{KarpathyCVPR14}), often rely on user-attached tags, resulting in noisy labels. As such, self-supervised learning has become a frequently used approach for learning video representations (\cite{Vondrick_2018_ECCV, Agrawal_2015_ICCV, DBLP:journals/corr/WangG15a}). Transformer-based cross-modal models tackling video and language are relatively newly emerging research topic, and have displayed varying approaches with more task-oriented principles. While many models adopted approaches employed in the models for images, for example the BERT-inspired pre-training objectives, there are also models that opt to diverge, as we shall see below.

When aligning video and language, where language supervision is extracted from the corresponding audio of the unannotated raw videos, a frequent problem is that the semantic of the language and video may not align. For example, the speaker may be talking about cars, where the video shows the speaker himself. (\cite{Sun_2019_ICCV}) notes that cooking videos have high probability of visual and linguistic semantics temporally well-aligned, and exploit such properties to examine video caption generation and next frame prediction with their VideoBERT. In order to extend BERT's pre-training objectives to video domain, they obtain visual tokens using hierarchical vector quantization to video features. On top of the masked language modeling and masked visual token modeling, next-sentence classification is extended as alignment classification for visual and linguistic sentences. Even with cooking videos, however, the alignment may still be noisy, and they deal with this problem by concatenating neighboring sentences into a long single sentence, and varying the subsampling rate of video tokens, making the model robust to variations in video speed.

Similarly to VideoBERT, CBT (\cite{sun2020learning}) employs sliding window approach to extract visual tokens using S3D network (\cite{xie2018rethinking}). Visual and linguistic tokens are concatenated and passed to a shallow 1-layer cross-modal transformer, where the mutual information between the two is computed. For pre-training objectives, masked language modeling and masked visual token modeling are employed, where HowTo100M dataset (\cite{miech19howto100m}) is used, rather than crawled videos from YouTube as is the case with VideoBERT. Also, unlike VideoBERT where the features are clustered with vector quantization and the task is to estimate to which cluster the masked feature belongs to, CBT attempts to directly regress the masked features. While the models above rely on densely sampled visual tokens, ClipBERT (\cite{lei2021more}) notably demonstrates that sparse visual tokens can be sufficient for video representation.

Novel pre-training objectives that are specific to video and language domain have also appeared. For example, ActBERT (\cite{Zhu_2020_CVPR}) proposes masked action classification, where the objective is to predict the action label given the linguistic and object features, whereas HERO (\cite{li2020hero}) proposes frame order modeling. VideoAsMT (\cite{korbar2020video}) and UniVL (\cite{luo2020univl}) also incorporate generative modeling approach, by proposing to regard each modality as a translation of each other.

Models that do not explicitly employ BERT-inspired approaches have also appeared. (\cite{gabeur2020mmt}) proposes multi-modal transformer calibrated for video retrieval task. Rather than directly extending pre-training objectives of BERT, they rely on bidirectional max-margin ranking, enforcing high similarity between video and captions. Notably, they extract video features by the combination of pre-trained models from 7 different domains, which differentiates them from other models that generally employ one pre-trained model for each modality. Similarly, Cooperative hierarchical transformer (COOT) (\cite{ging2020coot}) proposes cross-modal cycle-consistency loss to enforce semantic alignment, on top of alignment losses from (\cite{zhang2018crossmodal}), rather than simply extending pre-training objectives of BERT. (\cite{Zhou_2018_CVPR}) also proposes to use masked transformer dense video captioning task, but simply uses it as a captioning decoder based on inputs from visual encoder and proposal decoder, without explicit pre-training objective designed for cross-modal learning.

As we have seen, while there are only a limited number of transformer-based models dealing with video and language, they tend to be specific to task and domain, taking varying approaches. Many models have employed approaches from cross-modal models for image and language, but it may be still early to capture a definite common ground for these models, since much more remains to be explored yet. Table~\ref{table:video_caption} summarizes the models' performances on video captioning task, one of the frequently visited tasks in video and language cross-modal domain.

\subsection{Network Architecture}
\label{sec3:2}

\begin{figure*}
\centering     
\subfigure[Single-stream]{\label{fig:a}\includegraphics[width=43.5mm]{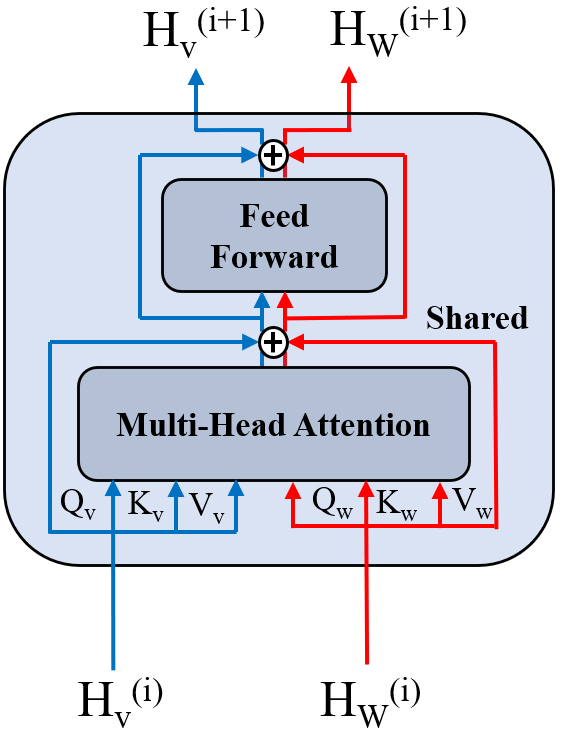}}
\subfigure[Two-stream]{\label{fig:b}\includegraphics[width=60mm]{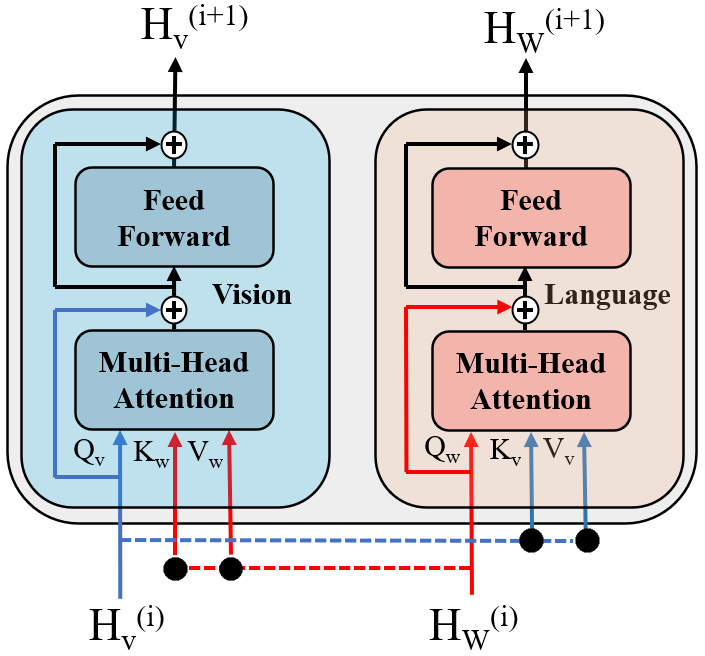}}
\caption{Comparison of single-stream and two-stream cross-modal transformer blocks. $H^{(i)}_v$ and $H^{(i)}_w$ refer to embedding of visual and word tokens respectively, output by \textit{i}-th layer.}
\label{fig:coattention}
\vspace{-3mm}
\end{figure*}

Network architecture for cross-modal embedding using transformer can be roughly divided into two categories; single-stream models, where the transformer block is modality-specific, and two-stream models, where the inputs to each transformer block are inter-modal. (Figure~\ref{fig:coattention})

ViLBERT (\cite{NEURIPS2019_c74d97b0}) is a representative two-stream model, in which they proposed co-attention transformer mechanism, with input keys and values of one modality being passed as inputs to the transformer block of another modality. In other words, keys and values for the language are input to transformer block for the vision part, and vice versa. Since the queries for each modality still go into the corresponding modalities, the transformer block ends up learning to embed features for each modality conditioned on the other modality. In order to tokenize the vision, they extract image regions using Faster R-CNN with ResNet (\cite{7780459}) backbone, while also employing 5-dimensional spatial location vector. LXMERT (\cite{tan-bansal-2019-lxmert}) takes a similar approach, where, after embedding each modality and encoding them with transformer separately, cross-modality encoder is applied, for which query vector $h^{k}_i$ from one modality and context vector $v^{k}_j$ from another modality are the inputs, with $k$ being the number of respective single-modality encoders. DeVLBERT (\cite{Zhang_2020}) and (\cite{kervadec2019weak}) also follow ViLBERT by exchanging queries, or equivalently keys and values, for each modality. SemVLP (\cite{li2021semvlp}) notably performs two-stream cross-modal learning only at the upper parts of the blocks, in order to learn high-level semantic alignment. ERNIE-ViL (\cite{yu2020ernievil}), while mostly following ViLBERT, incorporates scene graph approach, where the input text is parsed into the nodes of objects, relations, and attributes that the text mentions. Scene graph representation is shown to improve the semantic alignment between vision and language, and as the authors note, hints at the possibility of extending cross-modal tasks to graph neural networks. 

Single-stream cross-modal learning is fairly intuitive and simple, as the conventional transformer blocks can be extended in a straightforward manner, without architectural modifications, for concatenated inputs. Unicoder-VL (\cite{Li_Duan_Fang_Gong_Jiang_2020}), UNITER (\cite{chen2020uniter}), and VisualBERT (\cite{li2019visualbert}) are some of the notable examples of single-stream cross-modal models.

Notably, nearly all models perform early fusion of language and vision embeddings, in which they are concatenated prior to being fed to cross-modal transformer blocks. In fact, B2T2 (\cite{alberti-etal-2019-fusion}), designed specifically for VQA and VCR tasks, modifies dual encoder (\cite{wu2017starspace, DBLP:journals/corr/abs-1811-08008}) and compares performing early fusion with text and bounding boxes to late fusion with the feature from the entire image, and reports that early fusion outperforms the late fusion.

MiniVLM (\cite{wang2020minivlm}) aims at building a light-weight cross-modal model, and unlike the models that employed Fast R-CNN or Faster R-CNN to extract visual features, they propose two-stage efficient feature extractor (TEE), consisting of EfficientNet (\cite{pmlr-v97-tan19a}) and compact BERT fusion model. In particular, the visual feature extraction consists of EfficientNet with bidirectional feature pyramid network as in EfficinetDet (\cite{tan2020efficientdet}), followed by non-maximum suppression. MiniVLM is also unique in that they propose a triplet input to cross-modal learning block, consisting of visual features, tokenized sentences, and tokenized object names. The motivation is that inputting object labels explicitly may help further enforce the learning of dependency between objects and corresponding text. Similar approach is also employed by OSCAR (\cite{li2020oscar}), which inputs object tags along with word tokens and region features.

Apart from the cross-modal learning mechanism and input format, positional embedding is another source of variation among the models. Unicoder-VL (\cite{Li_Duan_Fang_Gong_Jiang_2020}) extracts image regions with Faster R-CNN along with 5-d vector, but they notably use the same position embedding for all image regions, rather than random permutations. In VisualBERT (\cite{li2019visualbert}) position embedding for each visual token is matched to the corresponding input token, whenever the alignments between image regions and the input tokens are available. 

VL-BERT (\cite{Su2020VL-BERT:}) proposes a format where each input element consists of 4 different types of embeddings, namely token embedding, visual feature embedding, segment embedding, and sequence position embedding. Most notable is visual feature embedding, which is a combination of visual appearance feature, corresponding to the output of fully-concatenated features from Fast R-CNN for the image region, and visual geometry embedding, a 4-d vector for coordinates of each corner of the region. Note that, for linguistic tokens, the entire image is used for visual feature embedding. Sequence position embedding for image tokens can be randomized. 

Both single-stream and two-stream have appeared in video and language domain as well. For example, ActBERT proposes tangled transformer, where each of query, key, and value come from different modalities, while UniVL employs cross encoder, where language and video encodings are combined along the dimension of sequence. On the other hand, E2vid (\cite{huang-etal-2020-multimodal}) relies on separate-modality architecture, and only partially employs cross-modal co-attention transformer.

In summary, we can make a few observations that are common in most models. While some models explicitly come up with two-stream cross-attention scheme by exchanging inputs to attention heads, many models do not explicitly design cross-attention scheme and simply rely on pre-training objectives for learning cross-modal dependency. Whether single-stream or two-stream, images are tokenized as regions without exception, although the entire image may replace regions in some models for position embedding. Also, regions are mostly accompanied by low-dimensional spatial vectors containing their coordinate information. Features are extracted with pre-trained object detection models rather than classification models, except models like MiniVLM, which employs EfficientNet backbone and further trains the detection model. Attaching position embedding to image regions is one source of variations among the models. Many models employ coordinate-based ordering, while some models simply use random permutation, or the same position embedding for all visual tokens.

\section{Visual Representation with Transformer}
\label{sec:4}
So far, most works tackling cross-modal tasks with transformer architecture have primarily applied it to linguistic representation, with certain schemes to embed it together with visual representation, mostly by tokenizing the visual representation in some ways. Such tokenization of visual representations has for the most part relied on pre-trained convolutional neural networks. For example, ViLBERT (\cite{NEURIPS2019_c74d97b0}) relies on Faster R-CNN to extract image regions as tokens. However, some of the recent researches have suggested that even convolution may be safely replaced by transformer architecture, which may arguably imply that more fundamentally different changes may be possible with obtaining visual representations. While these works do not explicitly deal with cross-modal tasks, they suggest critical implications as to the prospects of cross-modal tasks, as will be discussed in Sec~\ref{sec:5}. We thus briefly introduce recent important works on vision representation with transformers in this section.

(\cite{dosovitskiy2020image}) proposes Vision Transformer (ViT) and suggests that pure transformers can achieve comparable performance on image classification tasks. While closely following the original transformer architecture, they split images into patches, feed the sequence of linear embeddings of these patches as input to transformer, so that 2D images are represented as a sequence of flattened patches. In a similar spirit as masked language modeling in BERT, they perform  \textit{masked patch prediction}, in which the model is trained to predict the mean 3bit color for each patch. In particular, their ablation study with variations in dataset size shows that pre-training with the largest dataset, JFT-300M (\cite{Sun_2017_ICCV}), resulted in better performance, implicating that, as in language domain, training a transformer-based model with a massively large amount of image data can lead to models with outstanding performances. Pyramid vision transformer (\cite{DBLP:journals/corr/abs-2102-12122}) expands upon vision transformer and demonstrates that convolution-free models can be extended to a wider range of computer vision tasks, including object detection, semantic segmentation, and instance segmentation. VidTr (\cite{DBLP:journals/corr/abs-2104-11746}) further shows that videos can also be handled without using convolutions.

iGPT (\cite{pmlr-v119-chen20s}) proposes another direction for applicability of transformer in vision. It proposes to train a sequence transformer for pixel prediction, first by reshaping the pixels into 1D sequence, then by performing pre-training objectives of auto-regressive next pixel prediction and masked pixel prediction. The model that is trained at the scale and architecture of GPT-2 turns out to outperform Wide ResNet on CIFAR-10 pixel prediction task. This result implies a promising research direction for image generation and image enhancement with transformer. In fact, Image Transformer (\cite{parmar2018image}) and image processing transformer (IPT) (\cite{chen2020pretrained}) have demonstrated that transformer-based models can outperform conventional models in various tasks, such as image super-resolution and de-noising.

On top of the works introduced above, transformer architecture has already started to tackle other major computer vision tasks, mostly in conjunction with CNNs. For example, DETR (\cite{carion2020endtoend}) employs transformer encoder and decoder on top of CNN backbone for object detection, whereas Max-DeepLab (\cite{wang2020maxdeeplab}) performs panoptic segmentation by dual architecture of CNN and mask transformer, just to name a few. For further details on the application of transformer for general computer vision tasks, we refer the readers to two survey papers (\cite{han2021survey, khan2021transformers}).

\section{Prospects}
\label{sec:5}

One of the biggest concerns raised about the transformer-based language models is its massively large scale of training a model which is inevitably accompanied by prohibitive financial costs to afford such training procedure that have skyrocketed. (\cite{sharir2020cost}) has estimated that it would cost \$80k to \$1.6m to train a a 1.5 billion parameter model, and now we have models with over 100 billion parameters (\cite{NEURIPS2020_1457c0d6}), further aggravating such concerns regarding costs. Moreover, such tendency has also proven true for vision and cross-modal tasks; as noted in Section~\ref{sec:4}, training a model with larger image data led to noticeable performance boosts, and as Table~\ref{table:performance} shows, nearly all cross-modal models show in their ablation study that they have benefited immensely from training with larger models. This rapid inflation in training costs has raised concerns, as the affordability is limited to only a handful of large corporations. 

While potential alternatives may require a community-wise discussion, one of the most unambiguous directions, as (\cite{sharir2020cost}) points out, would be to develop more efficient network architectures; in fact, research efforts focusing on lessening the computational burden of transformer-based models have subsequently appeared in each modality. For example, in language domain, DistillBERT (\cite{sanh2020distilbert}) reduces the number of parameters by 40\% from BERT, while retaining nearly identical performance, with its knowledge distillation integrated to the loss function. TinyBERT (\cite{jiao2020tinybert}) similarly proposes to leverage knowledge distillation, and reports comparable performances with up to 7.5 times smaller model. In vision domain, data-efficient image transformer (DeiT) (\cite{touvron2020training}), built upon ViT, has been introduced, with an introduction of distillation token that utilizes student-teacher strategy through attention, and demonstrates a performance comparable to CNNs, while trained on ImageNet only. Reformer (\cite{Kitaev2020Reformer:}) is also an attempt aimed at improving the efficiency of transformer, and approaches the goal with locality-sensitive hashing and reversible residual layers, reducing its complexity from exponential to logarithmic. Sparse transformers (\cite{DBLP:journals/corr/abs-1904-10509}) and switch transformers (\cite{fedus2021switch}) also strive to diminish computational complexity with sparse attention mechanism. These results in respective domain and general transformer architecture bring high anticipation towards subsequent light-weighted transformer-based models in cross-modal tasks.

As shown in Sec~\ref{sec:4}, transformer architecture has been shown to be effective not only in language, but also in vision, even when no convolutional neural networks are involved. From such results, we speculate that a cross-modal model solely based on transformer architecture, in which both language and vision representations are acquired solely through transformer without convolution and recurrent neural networks, is imminent. Such possibility is important because the resulting architectural integrity is likely to lead to a development of hardware optimized for transformer architecture, as recent works claim transformers' superiority over CNNs in terms of computational efficiency. For example, (\cite{dosovitskiy2020image}) shows that ViT is up to 4 times more memory-efficient than ResNets. If such memory efficiency is coupled with distillation models, the cost-performance issues described above will be further alleviated to a substantial extent, and architectural transitions in much wider parts of deep learning are likely to follow. In fact, ViLT (\cite{kim2021vilt}) and UniT (\cite{DBLP:journals/corr/abs-2102-10772}) have been proposed to demonstrate that convolution-free model based on transformer for both modalities can accomplish comparable or better performances, while being up to 60 times faster. VATT (\cite{DBLP:journals/corr/abs-2104-11178}) goes a step further to include raw video and audio on top of text, with convolution-free transformer model. It will also be of our interest to see how the pre-training objectives evolve with the emergence of transformer-exclusive models. We expect more cross-modal models based on transformer architecture for both modalities to follow in near future, and predict that they are likely to play an important role in the transition of deep learning as described above.



As shown in Sec~\ref{sec:4}, iGPT (\cite{pmlr-v119-chen20s}) demonstrated that transformer can be employed to generate images. Extending this accomplishment to cross-modal domain, we encounter the task of image synthesis from text description. While image generation from text has been a long-time aspiration in machine learning community, its success has been limited; for example, to specific domains, such as birds or flowers (\cite{reed2016generative, Zhang_2018_CVPR}). However, Dall-e (\cite{ramesh2021zeroshot}) used GPT-3-based language model to demonstrate image synthesis from text with a wide variety, ranging from realistic images to illustrations, encompassing geographic and temporal knowledge. Along with text tokens, they tokenized images using discrete latent codes learned via discrete VAE (\cite{Kingma2014, pmlr-v32-rezende14}). StyleCLIP (\cite{DBLP:journals/corr/abs-2103-17249}) also demonstrates that image style can be manipulated with text input by leveraging contrastive language-image pre-training (\cite{radford2021learning}) on StyleGAN-based image generation pipeline (\cite{8953766, Karras_2020_CVPR}). This line of work is particularly intriguing since most cross-modal works employing transformer architecture have tackled the tasks in which images or videos are the inputs along with text, and the text or a label is the output. In that sense, image synthesis from text is a task in its reversed direction, and along with the combination with existing generative models, it is expected to be a highly promising future research direction.

\section{Conclusion}
\label{sec:6}
In this paper, we reviewed the recent trends of transformer-based models for cross-modal tasks with language and vision, with emphasis on pre-training scheme and network architecture. As in language domain, the performance for cross-modal models turns out to strongly depend on the aspects such as model size, dataset size, and the pre-training objectives, reinforcing the recent trend of bigger models and more data for better performance witnessed by the likes of GPT-3. We have also shown that, with other things being equal, pre-training with datasets annotated manually by humans is more efficient in terms of the necessary dataset size than pre-training with datasets whose annotations are obtained under weak supervision via crawling, although the discrepancy can be overcome with a sufficient amount of data. We also observed that pre-training directly with datasets used for target task can help achieve better performance on that particular target task.

We also introduced the works that applied transformer architecture for vision representation, and discussed potential prospects with regards to transformer-based cross-modal models, such as the potential of transformer-exclusive models, data-efficient models, and the models for generative tasks. As the topics discussed in this paper are still relatively in their early stages, we hope this paper serves as a useful reference that summarizes the first phase of the topic that will potentially grow increasingly important in deep learning field.



{\small
\bibliographystyle{spbasic}
\bibliography{ijcv}
}

\end{document}

%% file: macro.tex
\usepackage{xspace}
\usepackage{color}
\usepackage{multirow}
\usepackage{adjustbox}
\usepackage{subfigure}
\usepackage{paralist} 
\usepackage{enumitem} 
\usepackage{booktabs} 
\usepackage{array}

\graphicspath{{figures/}}

\usepackage[pagebackref=true,breaklinks=true,letterpaper=true,colorlinks,citecolor=blue,linkcolor=blue,bookmarks=false]{hyperref}

\long\def\ignorethis#1{}

\definecolor{orange}{rgb}{1,0.55,0}
\definecolor{gray}{rgb}{0.35,0.35,0.35}
\definecolor{MyBlue}{rgb}{0,0.2,0.8}
\definecolor{MyRed}{rgb}{0.8,0.2,0}
\definecolor{MyGreen}{rgb}{0.0,0.5,0.1}
\definecolor{MyGray}{rgb}{0.4,0.4,0.4}

\newlength\paramargin
\newlength\figmargin
\newlength\secmargin

\newcolumntype{L}[1]{>{\raggedright\let\newline\\\arraybackslash\hspace{0pt}}m{#1}}
\newcolumntype{C}[1]{>{\centering\let\newline\\\arraybackslash\hspace{0pt}}m{#1}}
\newcolumntype{R}[1]{>{\raggedleft\let\newline\\\arraybackslash\hspace{0pt}}m{#1}}

